\documentclass{article}

\usepackage{microtype}
\usepackage{graphicx}
\usepackage{subfigure}
\usepackage{booktabs} % for professional tables
\usepackage{amsmath}

\usepackage{hyperref}

\usepackage[accepted]{icml2019}

\icmltitlerunning{Progress Extrapolating Algorithmic Learning to Arbitrary Sequence Lengths}

\begin{document}

\twocolumn[
\icmltitle{Progress Extrapolating Algorithmic Learning to \\ Arbitrary Sequence Lengths}



\icmlsetsymbol{equal}{*}

\begin{icmlauthorlist}
\icmlauthor{Andreas Robinson}{equal,to}
\end{icmlauthorlist}

\icmlaffiliation{to}{NA}

\icmlcorrespondingauthor{Andreas Robinson}{robi0258@umn.edu}

\icmlkeywords{Machine Learning, ICML}

\vskip 0.3in
]



\printAffiliationsAndNotice{}  % leave blank if no need to mention equal contribution

\begin{abstract}
    Recent neural network models for algorithmic tasks have led to significant improvements in
    extrapolation to sequences much longer than training, but it remains an outstanding problem
    that the performance still degrades for very long or adversarial sequences.  We present
    alternative architectures and loss-terms to address these issues, and our testing of these
    approaches has not detected any remaining extrapolation errors within memory constraints.
    We focus on linear time algorithmic tasks including copy, parentheses parsing, and binary
    addition.  First, activation binning was used to discretize the trained network in order to avoid
    computational drift from continuous operations, and a binning-based digital loss term was
    added to encourage discretizable representations.  In addition, a localized differentiable
    memory (LDM) architecture, in contrast to distributed memory access, addressed remaining
    extrapolation errors and avoided unbounded growth of internal computational states.  Previous
    work has found that algorithmic extrapolation issues can also be alleviated with approaches
    relying on program traces, but the current effort does not rely on such traces.
\end{abstract}

\section{Introduction}
\label{introduction}

In recent years there has been substantial progress applying neural networks to
algorithmic problems such as addition, multiplication, and sorting.
These approaches, including Neural Turing Machines (NTMs) and Neural GPUs,
have shown large improvements in extrapolation to sequence lengths much longer than those encountered during
training, when compared to previous approaches such as LSTMs or LSTMs with
attention \cite{graves14, kaiser16}.

However, even with these improvements it remains an outstanding problem that neural network accuracies
often degrade with increasing sequence length or with adversarial sequences \cite{price16, freivalds18}.  In this paper, we experimented with three approaches to alleviate this issue:
binning-based discretization, incorporating a digital loss term, and using localized differentiable memory (LDM).
With these approaches the accuracies are 100\% on all cases that we have tested, including randomly generated
as well as adversarial examples.

First we experimented with binning the activations during inference, as well as adding a loss-term to
encourage the activations towards a small number of discrete bins.  This approach was motivated by the issue
that continuous (as opposed to discrete) operations risk gradually accumulating small errors across long sequences,
potentially leading to extrapolation errors.  We initially tried discretizing the activations
to 0 or 1, but in practice found that at least three discrete activation bins were needed for 100\% accuracy.

This binning approach serves a somewhat related purpose as the tanh cutoff approach from the Neural GPU paper
\cite{kaiser16} as well as the hard tanh used in \citet{freivalds18}, since those approaches do
effectively cut off large
magnitudes to $\{-1,1\}$; however, for smaller intermediate values they continue to allow continuous variation,
whereas the current binning approach modifies the full range of activations to a discrete set of
options and so more fully digitizes the computation.

The above binning approach, with a Neural GPU as the baseline network, was sufficient for some tasks
(e.g. summation), but continued to show small extrapolation errors in other tasks (parentheses parsing). To
address these remaining errors, we experimented with an alternative network model which relies
on localized differentiable memory (LDM). With this localized (as opposed to distributed) memory access, combined
with binning and digital loss, we obtained 100\% extrapolation on the remaining tests.

A potential explanation for improved extrapolation with localized memory access is that
(empirically) the number of discretized (binned) internal memory states steadily increased with input length
for the discretized Neural GPU model.  Whereas the number of states remained constant with input length
for the localized differentiable memory (LDM) based network.  Note that an algorithm that relies on indefinitely increasing
internal states cannot extrapolate beyond a threshold length, since only a finite set of discrete internal states
are available in a given network structure with finite precision (not including memory).

\begin{table*}[t]
    \centering
    \caption{Classification accuracies when extrapolating to sequences of length 900,
    from training length \textless= 20. Results shown for Neural GPUs and
    Localized-Differentiable-Memory (LDM) Networks, with and without binning and digital-bin loss.}
    \label{extrapolation-results}
    \vskip 0.15in
    \begin{center}
    \begin{small}
    \begin{sc}
    \begin{tabular}{lcccr}
    \toprule
    Algorithm & Copy Task & Sum & Sum Advers. & Par. Parsing \\
    \midrule
    neural-gpu   & 100.0\% & 97\% & 98.5\% & 96.99\% \\
    neural-gpu, binned & 100.0\% & 100.0\% & 100.0\% & 99.0\%\\
    ldm     & 58\% & 100.0\% & 100.0\% & 100\% \\
    ldm, binned  & 58\%  & 100.0\% & 100.0\% & 100.0\%  \\
    ldm, binned, dig-loss  & 100.0\% & 100.0\% & 100.0\% & 100.0\% \\
    \bottomrule
    \end{tabular}
    \end{sc}
    \end{small}
    \end{center}
    \vskip -0.1in
\end{table*}

\section{Related Work}

Algorithmic learning work builds on a large body of earlier research in the area of program induction.
This includes: \citet{liang13, nordin97, wineberg94, solomonoff64, holland92, gomez08, goldberg89}.
More recent work has emphasized neural network based
approaches, trainable end-to-end with gradient-based search
\cite{zaremba14, kaiser16, graves14, graves16, kurach16, andrychowicz16, deghani19}.  Approaches have been developed
to support external differentiable memory decoupled from computational weights \cite{graves14} as well as
automatically learned iteration counts \cite{graves16_act}.  These types of approaches can support turing completeness
\cite{deghani19} even with finite numerical precision. In practice much of the focus has been on learning linear
time algorithms, though Neural GPUs (for example) have succeeded in learning polynomial time algorithms end-to-end,
including binary and decimal multiplication \cite{kaiser16}.  These techniques have
also shown promise when applied to non-algorithmic tasks including translation \cite{deghani19} and
language-based reasoning \cite{graves16, deghani19}.

While many algorithmic tasks can be learned to some degree using more traditional recurrent networks
such as LSTMs, they are generally less effective at extrapolating to lengths much longer than the training
sequences \cite{graves14, kaiser16}.  However, even with state-of-the-art approaches, results still generally
degrade with increasing sequence length \cite{price16, freivalds18}.  Also, adversarial sequences can continue to
cause problems even when extrapolation is quite effective on random samples, e.g. digit-summation examples designed
to require carrying digits across long sequences \cite{price16}.  Somewhat related to our localized memory
approach, \citet{rae16} developed methods
for sparse differentiable memory; however, attention was still distributed across memory and the primary
improvement was
computational/resource efficiency (via sparse access) not improved extrapolation.

An alternative approach to algorithmic learning is to allow the learner access to program traces, providing
implementation details regarding the intended solution in addition to the usual input/output examples.  Our
results in this paper are focused on the case where traces are not available; however, trace-based approaches (e.g.
Neural Programmer-Interpreters) have demonstrated effectiveness on challenging algorithmic problems \cite{reed16},
and when combined with recursive
function calls they have even allowed extrapolation to arbitrary sequence lengths \cite{cai17}.

\section{Model Description}

\subsection{Activation Binning}\label{activation_binning}

One hypothesis for gradual degradation in accuracy with sequence length is that small
errors in
continuous neural network operations may accumulate across large numbers of iterations and operations.
A simple
approach one might try to prevent this would be to round all activations during inference,
e.g. rounding sigmoids to $\{0,1\}$, in order to make the computations more digital and avoid drift.
In the terminology of the current paper, this corresponds to having two activation bins; however, in practice
 more than
two are typically required to maintain the same validation accuracy as the pre-binned network.

So to discretize the activations, we create $N_b$ equally spaced bins, and round network activations
to the closest bin value.  In general, the bin values are given by:
\begin{align}
    bins &= \{min, min+s,..., max\} \\
    s &= (max - min) / (N_b - 1)
\end{align}
For sigmoid and softmax $\{min, max\}$ corresponds to $\{0,1\}$, whereas for tanh it is given by
$\{-1, 1\}$.  The number of bins $N_b$ is determined by starting with $2$ and then incrementing until
the binned
validation accuracy is at least as high as the unbinned accuracy; this is generally
100\% in this
paper, since the validation lengths are much shorter than the test lengths used to assess extrapolation.

This binning procedure is performed during the inference step, but not during training.
Also, not all activations
in the network are binned; in particular, the outputs
of each network iteration are binned, but not outputs of intermediate calculations used to
compute an iteration.  So for example when applying binning to the Neural GPU, the following are
binned: the tanh input embeddings, the tanh outputs after each recurrent network iteration, and the
final softmax output values; whereas the sigmoid gate values used within each iteration are not
binned.

In this paper, activation binning is applied both to a standard Neural GPU, as well as to
the localized differentiable memory (LDM) network described in \ref{ldm_memory}.

\subsection{Digital Loss}\label{digital_loss}

In many cases the number of required activation bins to maintain validation accuracy
has been quite small (2-5 bins).  However, for tasks/networks where large numbers of bins
are required,
e.g. due to the trained network relying on high precision in the activation values,
 the binning approach above can fail to have much impact on extrapolation.
 To address this issue, we added an additional term
 to the loss, in order to encourage the network activations to match a small target number of
 bins $N_t$.  This approach relies on the same subset of activations used by the activation
 binning in \ref{activation_binning}, which we can label ${a_1,...,a_c}$, where $c$ is the
 number of binned activations, not the number of bins.  The binning-based
 digital loss term is then given by:
\begin{equation}
    l_d = \frac{1}{c}\sum_{i=1}^{c} min(\{|a_i - b| : b \in bins\})
\end{equation}
where the set of equally spaced scalar values in $bins$ is determined as per
\ref{activation_binning}.  Also, the number of bins is $|bins| = N_t = min(N_b, N_m)$,
where $N_b$ is the number of
bins determined in \ref{activation_binning} for a given network model and task, and
$N_m$ is a hyperparameter representing
the max number of bins to target with digital loss (typically $N_m = 5$).  The digital
loss is added to the baseline loss $l_b$ with a weight-factor $w_d$, so
$loss = w_d*l_d + (1-w_d)*l_b$.

\subsection{Localized Differentiable Memory}\label{ldm_memory}

Localized Differentiable Memory (LDM) allows a network to read/write from
localized memory locations with read/write heads during each iteration, in contrast to
approaches such as NTMs which read/write to a large range of weighted
memory locations at each iteration.  Strictly speaking the localized approach is not fully differentiable,
but almost-everywhere differentiable, with a finite number of non-differentiable locations within a given
memory range, analogous to ReLU activations.

An LDM network is organized similarly to a Neural Turing Machine (NTM), i.e. a controller neural network
is connected recurrently to
an external memory via read and write heads.  However, unlike an NTM, but similar to a standard Turing machine,
an LDM read/write head has a specific location at each iteration, as opposed to reading from a large distributed
range of memory locations; for LDMs, almost-everywhere differentiability is
maintained
by representing the head positions as decimal rather than integer values.  So for instance a head position
of 3.2 will read from an average of memory locations 3 and 4, with $20\%$ of the read weight favoring
the former.  Also, like a standard Turing machine the heads can only move at most 1 step in either direction
at each iteration; however, with an LDM network the position shifts can be any decimal distance in the
range $(-1,1)$.  Continuing the analogy to standard Turing machines, you can think of the
controller neural network as effectively
implementing a Turing machine transition table.  Unlike NTMs there is no attention mechanism
in the current LDM implementation, but that is a
topic for future work.

In the current implementation the
controller network is represented
by a fully connected feed-forward neural
network with one hidden layer.  The activation for the hidden layer is ReLU, and the output activation is
sigmoid.

The inputs to the neural network at each iteration are given by:
\begin{equation}
    \label{inputs-formula}
    x_1^{t},...,x_d^{t}, v_r^{t}, s_1^{t}, ..., s_n^{t}
\end{equation}
where the $x_i^t$ values represent the current external inputs from the $t^{\text{th}}$ element of the input
sequence $x$, where each element of the sequence is of dimension $d$.  So the full input matrix $x$ is of
dimension $l \times d$, where $l$ is the sequence length. For classification of binary
sequences (e.g. parentheses parsing) $d$ is 1, whereas for pairwise binary addition $d$ is 2. Next, $v_r^{t}$
represent the previous values read from memory.  And the $s_i^{t}$ values are the states given
from the previous network controller outputs.  On the first iteration these recurrent feedback values are initialized
to 0.

The network outputs at each recurrent iteration are given by:
\begin{equation}
    m_r^{t+1}, m_w^{t+1}, \delta_r^{t+1}, \delta_w^{t+1}, v_w^{t+1}, s_1^{t+1}, ..., s_n^{t+1}
\end{equation}
Where the first five outputs are head control parameters, and the $s_i$ values represent the current stored state,
with $n$ state variables.  These values correspond to the outputs after the $t^{\text{th}}$ iteration.  A sigmoid activation is used on all the outputs, though $\delta_r$ and
$\delta_w$ are
linearly adjusted to  $(-1, 1)$.  Intuitively, $m_r$ and $m_w$ are booleans indicating whether to move
the read and write heads, $\delta_r$ and $\delta_w$ represent the movement distance from $(-1,1)$,
and $v_w$ represents the current value to write to memory.  For classification tasks,
the $s_n$ value after the final iteration represents the output result; for sequence-to-sequence,
the output sequence consists of the values $s_n^{1},...,s_n^I$, where $I$ is the total iterations.  For tasks where
no outputs are generated until all the inputs have been passed in (e.g. "copy" task), the output sequences
are prepended with $0s$ corresponding to the sequence length, and the inputs pass $-1s$ for the remaining
iterations after the input sequence.

The positions of the read and write heads are updated at each network iteration, $t$, according to:
\begin{align}
    p_r^{t+1} &= (1 - m_r) p_r^{t} + m_r (p_r^{t} + \delta_r) \\
    p_w^{t+1} &= (1 - m_w) p_w^{t} + m_w (p_w^{t} + \delta_w) \\
    p_r^0 & = p_w^0 = \lfloor{N_m/2}\rfloor
\end{align}
where $N_m$ is the integer size of the external memory array.  Also, memory overflow
is avoided by wrapping the memory, e.g. after the above head positions are computed, the
$p_r^{t+1}$ value is updated to $(p_r^{t+1}\ fmod\ N_m)$, and similarly for $p_w^{t+1}$;
note that this is using the floating point modulo
operator
$\textit{fmod}$ so
that the decimal components of the head positions
are preserved after wrapping.

Since the head positions are given by decimal values, the read operation takes a weighted average
of the two integer-indexed memory locations on either side of the head, weighted by the decimal component of the head
position.  So the read values at each network iteration are given by
${v_r}$:
\begin{align}
    j &= \lfloor{p_r^{t+1}}\rfloor \\
    w_j &= 1 - (p_r^{t+1} - j) \\
    v_r^{t+1} &= w_j\textbf{M}_j^{t} + (1-w_j)\textbf{M}_{j+1}^{t}
\end{align}
where $\textbf{M}$ is the memory array, with integer indexes and stored values
 in the range $[0, 1]$.  The integer memory index $(j+1)$ is wrapped to 0 if it exceeds the
 memory size, to avoid memory overflow.

After each neural network iteration the memory array $\textbf{M}$ is updated at the two memory locations adjacent
to the current decimal write head location $p_w$, weighted by the decimal component of the head position:
\begin{align}
    k &= \lfloor{p_w^{t}}\rfloor \\
    w_k &= 1 - (p_w^{t} - k) \\
    \textbf{M}_k^{t+1} &= w_k v_w + (1 - w_k) \textbf{M}_k^{t} \\
    \textbf{M}_{k+1}^{t+1} &= (1 - w_k) v_w + w_k \textbf{M}_{k+1}^{t}
\end{align}
As with the read head above, the integer memory index $(k+1)$ is wrapped to 0 if it exceeds the
memory size, to avoid memory overflow.

\begin{figure}[ht]
    \vskip 0.2in
    \begin{center}
    \centerline{\includegraphics[width=\columnwidth]{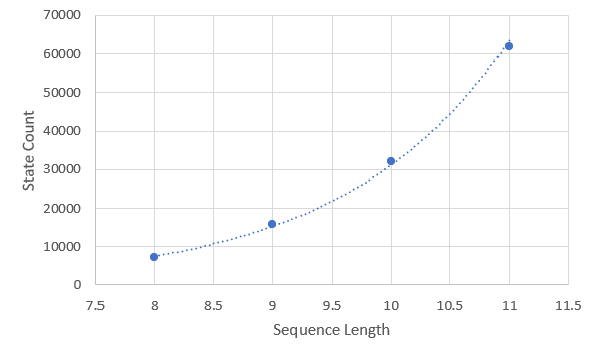}}
    \centerline{\includegraphics[width=\columnwidth]{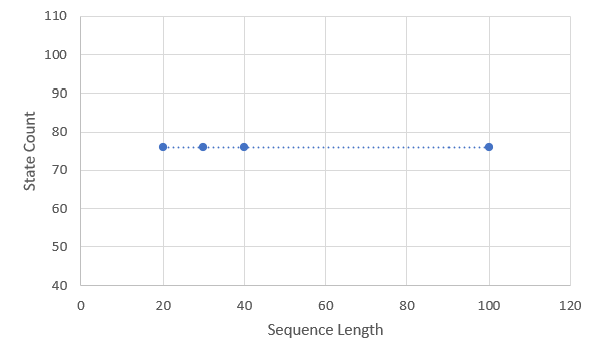}}
    \caption{The number of computational states plotted versus the input sequence length,
    for the parentheses parsing task.  The state counts are constant with
    input length for the binned LDM-network (bottom plot), whereas the binned Neural GPU fits an
    exponential growth curve with an $R^2$ of $.9988$ (top plot).}
    \label{fig-state-counts}
    \end{center}
    \vskip -0.2in
\end{figure}

\section{Experiments}

\subsection{Algorithmic Extrapolation}

Experiments were performed to assess algorithmic extrapolation to lengths much longer than training.
Training was on length 8-20 (validation 30) sequences and testing was
on length 900.  This test length was chosen since longer lengths gave out of memory errors
for some of the more memory intensive techniques (e.g. the 48 channel Neural gpu).

The algorithms tested were copy, binary sum, adversarial binary sum, and
parentheses parsing. In the copy task, the input sequence simply matches the output sequence.  In binary sum,
the
inputs and outputs are provided as aligned sequences with pairs of corresponding bits (i.e. matching
significant digits).  The adversarial summation task was taken from \citet{price16}, which
involves challenging summation samples that require carrying digits over many iterations.
For parenthesis parsing (Dyck words), roughly
half
the generated input sequences have properly matching left and right parentheses (represented by 0s and 1s) and
half the input input sequences are invalid sequences of parentheses.  The natural brute-force solution to the
parsing task
requires $O(N)$ memory.

The results are shown in Table~\ref{extrapolation-results}.  The extrapolation accuracies are 100\% across all the
target tasks when combining the three techniques from this paper: localized memory, activation
binning, and digital loss.  For comparison the baseline Neural GPU only achieved 100\% extrapolation
on the copy task.  Augmenting the Neural GPU with activation binning is sufficient for 100\%
extrapolation on all tasks except parenthesis parsing.

Also, the LDM-network alone, even without binning, extrapolates perfectly
on all tasks except the copy task.  Note that the copy task is more challenging for the LDM
approach than for Neural GPUs, since the former analyzes the input sequentially
(so it has to remember the earlier bits when copying them to the output), whereas the Neural GPU
processes the sequence in parallel and
simply has to learn the identity function on its input.  Additionally, binning alone was ineffective
for addressing the extrapolation issues with LDM-networks on the copy tasks, without
also applying digital loss; in particular, large numbers of
bins were required to get 100\% validation accuracy, but these fine-grained bins did not improve
extrapolation to the test set.  However, applying digital loss yielded 100\% validation with
just 5 bins, which then extrapolated perfectly on the test set.

Table \ref{extrapolation-results} only shows extrapolation to length 900, which was a lowest common
denominator length that was feasible for an apples-to-apples comparison
across all the learning approaches (given memory constraints).
 However, it is perhaps
worth noting that we have also not yet detected extrapolation errors in any informal
tests with longer sequence lengths, for the approach combinations that give 100\%
extrapolation in the results table.

We observed minimal downside from incorporating binning and digital loss,
relative to the extrapolation improvements; On the other hand, the LDM-network
was somewhat more challenging train compared to the baseline Neural GPU,
often requiring an iterative approach in which
the training sequence lengths were gradually increased during training, which was generally not necessary
for training
convergence with a Neural GPU.

\subsection{Counting Computational States}

Given the continuous operations of a typical neural network, it can be difficult to define or measure the
number of discrete computational
states for comparison with a standard algorithm or Turing machine.  However, once the network activations
are binned
into a discrete set of values we can count the number of distinct states used by the recurrent network.
In a Turing machine we would expect there to be a fixed total number of states available independent
of the input sequence length, but a potential risk with neural-network based approaches is that
they will learn
to use the states as a memory store, in which case the network will fail to extrapolate beyond the
maximum supported state count (which may be sufficient for all training lengths, but not necessarily test).
Note that when
measuring the states we exclude data that is stored
to extensible memory, as that can of course be unbounded.

For the binned LDM-network, the state count is measured based on the binned versions
of the network input values from equation \ref{inputs-formula}, corresponding to the
inputs at the current recurrent iteration.  These inputs include
the current external input, the previous outputs of the controller network, and the current value(s)
read from the memory head location.  The state count corresponds to the number of distinct values of
this state vector that occur when applying the network to arbitrary inputs of a given input length.
We then plot the number of states as a function of input length in figure \ref{fig-state-counts}.  Since
the state
as we've defined it includes both current memory read values in addition to the pure state variables, you can
think of this as analogous to counting the full set of entries in a Turing machine transition table, which
includes entries for each possible combination of read value and internal state.

Note that the
number of measured states increases as the number of input test sequences is increased (since each test sequence
does not require all the computational states), but we
increase the test set size until the state count reaches a plateaux, approximating the full
set of states needed to handle any input of the given length; these total (plateaued) state counts are the
values shown in the plot.

For the binned Neural GPU, the state count is measured with basically the same approach; however, there is a
somewhat less
clear division between memory and state in this case, since there is no separate external memory distinct
from the network
proper, as there is with the LDM-network (and NTM).  Rather, the hidden layers consist of $O(N)$ memory units, for
inputs of length $N$.  So distinct states are interpreted as the set of values that can be stored at a given
memory location (at the start of each iteration), since there is no additional state beyond what is stored in
the memory cells.  Note we focus on the distinct values at any given memory location, since we are interested in the
computational state not the unbounded set of full memory states; unlike memory storage this computational state
count should be bounded in order to support successful extrapolation to arbitrary lengths, since only a finite
number of states are available at a given memory location, assuming finite-precision activations.

Figure \ref{fig-state-counts} shows the number of computational states used by the networks as a function of the input
lengths, for the parentheses parsing task.  The top plot shows the binned Neural GPU, and the bottom plot shows the
LDM-network.  As can be seen from the plot the number of LDM-network states remained constant at 76 for each tested
sequence length.  Whereas for the binned Neural GPU, the number of states grew exponentially, with an $R^2$ of $.9988$
for an exponential curve fit.  We also performed state counts for the binary addition task, with similar results, namely
the LDM approach had constant state count, whereas the binned Neural GPU showed an exponential increase in state usage
with input length.

This suggests that one possible explanation for the improved extrapolation of the
LDM-network is that it is relying on extensible memory for storing data rather than depending on the computational
state; in particular, if the state count for the Neural GPU continues to grow with input length per
the measured trend, this places a limit on the max
sequence length that can be handled by the network, since finite precision limits the number of states that can be
tracked with a fixed set of channels.  While these results are suggestive, we can't yet rule out the possibility that
the states plateaux with larger inputs and aren't actually a factor in the observed extrapolation errors.

\section{Conclusion}

The results from table \ref{extrapolation-results} demonstrate that the approaches in this paper
are able to obtain 100\% extrapolation in cases which are challenging for state of the art approaches.
And  we have not encountered any cases where extrapolation was less than
100\% when
all three novel approaches were applied, within memory constraints.  It is perhaps not surprising that
binning and digital loss would
improve extrapolation, due to discretization avoiding the accumulation of errors from continuous
computation across long sequences; however, it is somewhat more uncertain why localized (LDM) memory provided
further improvements to extrapolation.  But the finding that the computational state usage remained much
more stable for this approach, versus increasing exponentially with binned Neural GPUs, suggests a potential
explanation, since finite precision places a limit on the available information that can be tracked with
state, as opposed to extensible memory. Another potential advantage of localized LDM memory is that it could
provide performance benefits for very large memory stores; on the other hand, it also has reduced parallelizability
in comparison to the Neural GPU.

One challenge with demonstrating a solution to the problem of algorithmic extrapolation is that it is
difficult to prove a negative and demonstrate that there is no sequence length at which extrapolation begins
to fail.  So one avenue for future research is to go beyond empirical tests and provide a mathematical
demonstration that the current approaches extrapolate to arbitrary lengths for some tasks
(or that they fail to do so), assuming unconstrained memory resources.  Also, the
target
tasks for this paper were focused on problems with no worse than $O(N)$ time/memory complexity. However,
end-to-end systems, e.g. Neural GPUs, can learn polynomial complexity task, so extending the current techniques
to those tasks would be a natural direction; it would be straightforward to test a binned Neural GPU (with
 digital loss)
on such tasks, but extending LDM-networks to polynomial problems, such as binary multiplication,
 could be more challenging;
one possible approach would be to incorporate an attention mechanism allowing some degree of random access
memory.  It would also be valuable to compare the approaches in this paper to a wider range of baseline
approaches including Neural Turing Machines (NTMs) and LSTMs.  Further investigation could also be warranted
to determine whether the exponential increase in computational states observed with binned Neural GPUs
is a causal factor in the observed extrapolation failures.

\bibliography{main}
\bibliographystyle{icml2019}

\end{document}